%% file: main.tex
\documentclass[11pt,a4paper]{article}

\usepackage[utf8]{inputenc}
\usepackage[T1]{fontenc}
\usepackage{amsmath,amssymb,amsfonts}
\usepackage{graphicx}
\usepackage{booktabs}
\usepackage{hyperref}
\usepackage{natbib}
\usepackage{xcolor}
\usepackage{microtype}
\usepackage{algorithm}
\usepackage{algorithmic}
\usepackage{subcaption}
\usepackage[margin=1in]{geometry}

\newcommand{\modelname}{IMU-1}
\newcommand{\tokensbudget}{72B}

\newcommand{\muP}{$\mu$P}

\title{\modelname{}: Sample-Efficient Pre-training of Small Language Models}

\author{
  George Grigorev \\
  Independent Researcher \\
  \texttt{deezchannel@gmail.com}
}

\date{}

\begin{document}

\maketitle

\begin{abstract}
We present \modelname{}, a 430M-parameter language model trained on \tokensbudget{} tokens that approaches the benchmark performance of models trained on 56$\times$ more data.
We describe a validated training recipe combining recent architectural interventions (QK-norm attention, per-head gating, value residuals, LayerNorm scaling) with optimization advances (NorMuon with cautious weight decay, \muP{} parametrization) and a three-stage training schedule with post-hoc checkpoint EMA.
We provide ablations for each component and release code, weights and data to enable reproduction: \url{https://huggingface.co/thepowerfuldeez/imu1_base}
\end{abstract}

\input{sections/introduction}
\input{sections/related_work}

\input{sections/method}
\input{sections/data}

\input{sections/eval}

\input{sections/results}

\input{sections/ablations}
\input{sections/repro}

\input{sections/limitations}

\bibliographystyle{plainnat}
\bibliography{refs}

\appendix
\input{sections/appendix}

\end{document}

%% file: sections/introduction.tex
\section{Introduction}
\label{sec:introduction}

The training of large language models (LLMs) has traditionally been characterized by ever-increasing compute budgets, with state-of-the-art models requiring trillions of tokens and thousands of GPU-hours.
However, recent work has demonstrated that careful attention to architecture, optimization, and data selection can substantially improve \emph{sample efficiency}---the quality achieved per training token~\cite{BenAllal2025SmolLM2,Li2024DataCompLM,EssentialAI2025Muon}.

This paper presents \modelname{}, a 430M-parameter language model trained on \tokensbudget{} tokens that surpasses SmolLM-360M (600B tokens) and approaches SmolLM2-360M (4T tokens) on standard benchmarks.
Rather than proposing novel techniques, we provide a validated recipe that combines multiple recent advances into a reproducible training pipeline:

\begin{itemize}
    \item \textbf{Architectural interventions:} QK-norm attention for stability~\cite{Henry2020QKNorm}, per-head gated attention for improved expressivity and attention sink mitigation~\cite{Qiu2025GatedAttention}, value residual connections for improved gradient flow~\cite{Zhou2024ValueResidual}, and LayerNorm scaling to address depth-related training pathologies.
    \item \textbf{Optimization advances:} NorMuon optimizer with Triton-accelerated orthogonalization~\cite{Li2025NorMuon}, cautious weight decay for selective regularization~\cite{Chen2025CWD}, and \muP{} parametrization for hyperparameter transfer~\cite{Yang2022TensorProgramsV}.
    \item \textbf{Training dynamics:} Warmup-Stable-Decay (WSD) learning rate schedules with flexible decay timing~\cite{Wen2024WSD}, and post-hoc exponential moving average (EMA) of checkpoints for improved final performance.
\end{itemize}

We examine each intervention through ablations (\S\ref{sec:ablations}), document our data mixture (\S\ref{sec:data}), and release code, weights, and configurations (\S\ref{sec:repro}).

\paragraph{Contributions.}
\begin{enumerate}
    \item A validated training recipe for sample-efficient small-LM pretraining, combining known architectural and optimization techniques.
    \item Ablations isolating the contribution of each intervention on a 70M proxy model.
    \item A complete, reproducible training framework with code, weights, data, and configurations.
\end{enumerate}

%% file: sections/related_work.tex
\section{Related Work}
\label{sec:related_work}

\paragraph{Sample-efficient pretraining.}
Recent work demonstrates that small models can achieve strong performance through careful data curation and training strategies.
SmolLM2~\cite{BenAllal2025SmolLM2} documents a data-centric approach using multi-stage mixtures of web text, math, code, and instruction data, achieving state-of-the-art results at the 360M--1.7B scale.
DataComp-LM~\cite{Li2024DataCompLM} provides controlled dataset experimentation with standardized evaluation, while RefinedWeb~\cite{Penedo2023RefinedWeb} demonstrates that careful web filtering alone can match curated datasets.
However, most prior work relies on cosine learning rate schedules and conventional optimizers.
Our work combines data-centric strategies with architectural and optimization advances that further improve sample efficiency.

\paragraph{Learning rate schedules.}
Cosine decay remains the dominant schedule for LLM pretraining, but it requires specifying the total compute budget in advance.
This constraint hinders practitioners with limited or uncertain compute, since the schedule cannot be adjusted based on intermediate evaluation results.
Warmup-Stable-Decay (WSD)~\cite{Wen2024WSD} addresses this by maintaining a constant learning rate during a ``stable'' phase, with decay applied at a configurable cutoff.
This flexibility enables mid-training decisions based on benchmark performance, and recent work shows WSD interacts favorably with data mixture changes during training~\cite{Luo2025LRDecayCurriculum}.

Checkpoint averaging improves final model quality when applied during the decay phase~\cite{Tian2025WSM}.
We adopt post-hoc EMA of weights.

\paragraph{Small language model architectures.}
Several 2025 releases target the sub-500M parameter range: Gemma-3-270M~\cite{Gemma32025} from Google, LFM2-350M~\cite{LFM22025} from Liquid AI using hybrid Mamba-attention, Granite-4.0-H-350M~\cite{Granite42025} from IBM also using hybrid architectures, and Baguettotron~\cite{Baguettotron2025} trained on synthetic data.
These models train on 10--14 trillion tokens.
In contrast, we demonstrate that combining multiple architectural interventions---QK-norm, per-head gating, value residuals, and LayerNorm scaling---with modern optimizers enables similar quality at 72B tokens.

\paragraph{Optimization advances.}
Muon-style orthogonalized updates~\cite{EssentialAI2025Muon} have emerged as a promising alternative to Adam for LLM training.
NorMuon~\cite{Li2025NorMuon} improves scalability through neuron-wise normalization, while Cautious Weight Decay~\cite{Chen2025CWD} provides selective regularization based on gradient-weight alignment.
Most prior small model work uses AdamW; we adopt NorMuon with cautious weight decay for improved training dynamics.

%% file: sections/method.tex
\section{Method}
\label{sec:method}

We present an approach to improving sample efficiency in language model pretraining through architectural and optimization interventions.
Our method addresses two complementary goals: (1) improving training stability to enable higher learning rates, and (2) improving gradient signal quality to accelerate convergence.
Each intervention is validated through controlled ablations (\S\ref{sec:ablations}), with combined improvements yielding 5.21\% relative reduction in training loss at matched tokens and iterations.

\subsection{Model Architecture}
\label{sec:base_arch}

\modelname{} is a 430M-parameter decoder-only transformer with the following configuration: $d_{\text{model}}=1152$, 30 layers, 18 attention heads, and grouped-query attention~\cite{Ainslie2023GQA} with 6 KV heads.
We use RMSNorm~\cite{Zhang2019RMSNorm} for normalization, SwiGLU~\cite{Shazeer2020GLU} for the feed-forward network, and rotary position embeddings (RoPE)~\cite{Su2021RoPE}.
The vocabulary size is 49,152 tokens using the SmolLM2-360M tokenizer~\cite{BenAllal2025SmolLM2}.

\paragraph{Ablation model.}
For ablations (\S\ref{sec:ablations}), we use a smaller 70M-parameter model ($d_{\text{model}}=768$, 8 layers, 12 heads, 4 KV heads) with a custom 32,768-token BPE tokenizer to enable rapid iteration.

\subsection{Architectural Interventions}
\label{sec:arch_interventions}

We apply four architectural modifications targeting attention stability, gradient flow, and depth scaling.
Our ablations (\S\ref{sec:arch_ablations}) show that while some interventions provide isolated benefits, their combined effect exceeds the sum of parts.

\subsubsection{QK-Norm Attention}
\label{sec:qknorm}

Standard scaled dot-product attention computes:
\begin{equation}
\text{Attention}(Q, K, V) = \text{softmax}\left(\frac{QK^\top}{\sqrt{d_h}}\right) V
\end{equation}
where $d_h$ is the head dimension.
As training progresses, the norms of $Q$ and $K$ can grow unboundedly, leading to large logits that saturate the softmax and produce high-kurtosis attention distributions~\cite{Dehghani2023ScalingViT,Wortsman2023StableTraining}.

Following \citet{Henry2020QKNorm}, we normalize queries and keys using RMS normalization before computing attention:
\begin{equation}
\text{Attention}(Q, K, V) = \text{softmax}\left(\gamma \cdot \frac{Q}{\text{RMS}(Q)} \cdot \frac{K^\top}{\text{RMS}(K)}\right) V
\label{eq:qknorm}
\end{equation}
where $\text{RMS}(x) = \sqrt{\frac{1}{d_h}\sum_i x_i^2}$ and $\gamma \in \mathbb{R}$ is a learnable gain parameter initialized to 1.
Normalization bounds attention logit magnitudes, preventing saturation regardless of $Q$ and $K$ scales.

We fuse QK-Norm into a custom FlashAttention~\cite{Dao2022FlashAttention} kernel for efficiency.

\subsubsection{Per-Head Gated Attention}
\label{sec:gating}

Transformers trained with causal attention develop ``attention sinks''---early tokens that accumulate disproportionate attention mass regardless of semantic relevance~\cite{Xiao2023StreamingLLM}.
This wastes model capacity and creates artifacts in streaming inference.

Following \citet{Qiu2025GatedAttention}, we introduce learnable per-head gates that modulate attention output:
\begin{equation}
\text{out}_h = 2 \cdot \sigma(g_h) \cdot \text{Attention}_h(Q, K, V)
\label{eq:gating}
\end{equation}
where $g \in \mathbb{R}^{n_h}$ is computed via projection $g = W_g x$ with $W_g \in \mathbb{R}^{d \times n_h}$, and $g_h$ is the scalar gate logit for head $h$.
The factor of 2 ensures the expected output magnitude matches standard attention when $g_h = 0$ (i.e., $\sigma(0) = 0.5$).

\subsubsection{Normalized Value Residual Learning}
\label{sec:val_residual}

Deep transformers suffer from gradient degradation in the value stream, as values pass through many layers of attention-weighted averaging before reaching the output~\cite{Zhou2024ValueResidual}.
This ``rank collapse'' reduces the effective expressivity of deep networks.

We extend the original value residual formulation with explicit normalization and disentangled scaling:
\begin{equation}
V^{(l)} = s \cdot \frac{\alpha_1 V^{(l)}_{\text{local}} + \alpha_2 V^{(1)}}{\sqrt{\alpha_1^2 + \alpha_2^2}}
\label{eq:val_residual}
\end{equation}
where $V^{(l)}_{\text{local}} = W_V^{(l)} x$ is the current layer's value projection and $V^{(1)}$ is the first layer's value projection.
We introduce three modifications: (1) normalization by $\sqrt{\alpha_1^2 + \alpha_2^2}$ preserves output magnitude as the residual weight changes, (2) a separate scaling factor $s$ decouples magnitude control from the mixing ratio, and (3) initialization $(s, \alpha_1, \alpha_2) = (1, 1, 0)$ starts from standard attention and learns the residual contribution.

\subsubsection{LayerNorm Scaling}
\label{sec:lns}

\citet{Sun2025CurseDepth} identify that deeper transformer layers contribute less to the output due to accumulated normalization effects---the ``curse of depth.''
Later layers' contributions are suppressed relative to earlier layers, limiting the effective depth of the network.

We apply depth-dependent scaling to LayerNorm outputs:
\begin{equation}
\text{LN}_l(x) = \frac{1}{\sqrt{l}} \cdot \text{LayerNorm}(x)
\label{eq:lns}
\end{equation}
where $l \in \{1, \ldots, L\}$ is the layer index.
This rebalances layer contributions, allowing deeper layers to have proportional impact on the output.

LayerNorm Scaling may become more impactful at greater depth.

\subsection{Optimization}
\label{sec:optimization}

Beyond architecture, we apply optimization techniques that improve gradient utilization and regularization.

\subsubsection{NorMuon Optimizer}
\label{sec:normuon}

AdamW applies element-wise adaptive learning rates, which can be suboptimal for weight matrices where parameters interact through matrix multiplication~\cite{EssentialAI2025Muon}.
Muon-style optimizers instead orthogonalize gradients before applying updates, ensuring updates remain on the Stiefel manifold of orthonormal matrices.

We use NorMuon~\cite{Li2025NorMuon}, which addresses norm imbalance after orthogonalization through neuron-wise normalization:
\begin{equation}
W_{t+1} = W_t - \eta \cdot \text{NeuronNorm}(\text{Ortho}(G_t))
\label{eq:normuon}
\end{equation}
where $G_t$ is the gradient, $\text{Ortho}(\cdot)$ applies Newton-Schulz iterations for orthogonalization, and $\text{NeuronNorm}(\cdot)$ normalizes each output neuron's update.

We implement orthogonalization using Polar Express constants~\cite{Amsel2025PolarExpress} with 7 Newton-Schulz steps, and use the Dion~\cite{Ahn2025Dion} Triton kernel for efficiency.
Muon tolerates higher learning rates than AdamW due to the implicit preconditioning from orthogonalization; we use $\eta = 0.0235$ for 2D parameters (weight matrices) and $\eta = 0.007$ for 1D parameters.
Following standard practice~\cite{EssentialAI2025Muon}, 1D parameters include: (1) all parameters with $\text{ndim} < 2$ (biases, LayerNorm gains), (2) embedding weights, and (3) the output projection (lm\_head).
These are optimized with AdamW while weight matrices use NorMuon.

\subsubsection{Cautious Weight Decay}
\label{sec:cwd}

Standard weight decay applies uniform $L_2$ regularization: $\Delta w = -\lambda w$.
However, decay can conflict with gradient updates, pushing weights in the opposite direction from the loss gradient~\cite{Chen2025CWD}.

Cautious Weight Decay (CWD) applies decay only when it aligns with the update direction:
\begin{equation}
\Delta w_{\text{CWD}} = \begin{cases}
-\lambda w & \text{if } \text{sign}(u) = \text{sign}(w) \\
0 & \text{otherwise}
\end{cases}
\label{eq:cwd}
\end{equation}
where $u = \text{NeuronNorm}(\text{Ortho}(G_t))$ is the orthogonalized update (not the raw gradient).
This prevents decay from counteracting beneficial updates while still regularizing weights that the optimizer is not actively using.

\subsubsection{Z-Loss Regularization}
\label{sec:zloss}

Following \citet{Chowdhery2022PaLM}, we apply auxiliary Z-loss to the logits:
\begin{equation}
\mathcal{L}_z = \lambda_z \cdot \log^2\left(\sum_i \exp(z_i)\right)
\label{eq:zloss}
\end{equation}
where $z$ is the logit vector and $\lambda_z = 10^{-4}$.
Z-loss penalizes large logit magnitudes, improving numerical stability in softmax computation without affecting the argmax.

\subsection{Learning Rate Schedule}
\label{sec:schedule}

We explore two learning rate schedules: cosine decay and Warmup-Stable-Decay (WSD).

\paragraph{Cosine schedule.}
Our baseline uses cosine decay~\cite{Loshchilov2017SGDR} from peak learning rate to $\eta_{\min} = 0.01 \cdot \eta_{\max}$ over the training horizon, with linear warmup for the first 1,000 steps.

\paragraph{WSD schedule.}
WSD~\cite{Wen2024WSD} maintains a stable learning rate after warmup, then decays in the final phase.
This enables checkpoint re-use: a model trained with WSD stable phase can be continued with additional data before decaying.
We use $1-\sqrt{t}$ decay profile and set stable-phase LR to 55\% of peak cosine LR based on preliminary experiments (0.013 for Muon 2D params, 0.0039 for 1D params).
Our ablations (\S\ref{sec:wsd_ablations}) evaluate decay fractions of 10--30\%.

\subsection{Implementation}
\label{sec:implementation}

We implement all techniques in PyTorch with custom Triton kernels for QK-Norm attention and NorMuon orthogonalization.
Training uses bfloat16 mixed precision with gradient checkpointing.
We apply \muP{} parametrization~\cite{Yang2022TensorProgramsV} for hyperparameter transfer across model scales.
Full implementation details and hyperparameters are provided in Appendix~\ref{sec:hyperparams}.

%% file: sections/data.tex
\section{Data}
\label{sec:data}

We use a mixture of web text (DCLM-edu, Cosmopedia), code (Stack-edu), math (FineMath), PDFs (FinePDFs), and wiki data (FineWiki), with progressively tighter quality filters in later stages~\cite{BenAllal2025SmolLM2}.

\subsection{Data sources and filtering}

\paragraph{Web data.}
DCLM-edu is filtered from DCLM~\cite{Li2024DataCompLM} using the FineWeb-edu classifier, which predicts educational quality on a 0--5 scale.
We use threshold $>2.75$ for Stage 1, increasing to $>3.2$ for Stage 2.
Cosmopedia-v2 synthetic textbooks are similarly filtered with the FineWeb-edu classifier (threshold $>2.0$ for Stage 1, $>2.3$ for Stage 2).

\paragraph{Code data.}
We follow The Stack V2 process~\cite{BenAllal2025SmolLM2}: scraping repositories from the latest available date, then filtering to Python-only files.
Educational quality is scored using the Stack-edu classifier with thresholds $>3.75$ (Stage 1) and $>4.1$ (Stage 2).

\paragraph{Math data.}
FineMath is filtered with threshold $>3.7$ (Stage 1) increasing to $>4.1$ (Stage 2).
We include MegaMath and other math-focused corpora in the mixture.

\paragraph{PDF and wiki data.}
FinePDFs are filtered with threshold $>1.75$ (Stage 1) to $>2.1$ (Stage 2).
We include FineWiki as a high-quality knowledge source.

\subsection{Quality thresholds by stage}

Table~\ref{tab:thresholds} summarizes filtering thresholds by stage.

\begin{table}[!htb]
\centering
\small
\begin{tabular}{lccc}
\toprule
\textbf{Source} & \textbf{Stage 1} & \textbf{Stage 2} & \textbf{Stage 3} \\
\midrule
DCLM-edu & $>2.75$ & $>3.2$ & --- \\
Cosmopedia & $>2.0$ & $>2.3$ & $>2.5$ \\
FineMath & $>3.7$ & $>4.1$ & $>4.25$ \\
Stack-edu & $>3.75$ & $>4.1$ & $>4.25$ \\
FinePDFs & $>1.75$ & $>2.1$ & $>2.4$ \\
\bottomrule
\end{tabular}
\caption{Educational score thresholds by training stage.}
\label{tab:thresholds}
\end{table}

\subsection{Stage 3: Midtrain}

Stage 3 additionally filters by document length ($>4000$ characters) and includes instruction-formatted data with chat templates.
We use synthetic reasoning data from Pleias SYNTH~\cite{Pleias2024Blogsynth} filtered with threshold $>1.65$ (Stage 1) to $>2.1$ (Stage 3).

\paragraph{DCLM-edu-boolq.}
We prepare a subset of DCLM-edu for Boolean question answering via model-based filtering.
An LLM identifies passages containing verifiable factual claims suitable for BoolQ/NLI-style tasks (Appendix~\ref{sec:boolq_prompt}), then we train a fastText classifier on these samples to label the full corpus.

\subsection{Training mixture by stage}

Table~\ref{tab:data_mixture} shows the available token pool for each training stage.
Training samples from these pools, consuming approximately 29B, 28B, and 14B tokens respectively (72B total).

\begin{table}[!htb]
\centering
\scriptsize
\begin{tabular}{lrrr}
\toprule
\textbf{Source} & \textbf{Stage 1 (B)} & \textbf{Stage 2 (B)} & \textbf{Stage 3 (B)} \\
\midrule
DCLM-edu & 27.0 & 5.1 & --- \\
FineWiki & 8.4 & 7.5 & --- \\
FinePDFs-edu & 8.0 & 5.3 & 4.2 \\
Python-edu & 12.0 & 1.4 & --- \\
FineMath & 5.0 & 2.0 & --- \\
MegaMath-web-pro & 5.0 & 5.1 & 1.3 \\
Zyda-2 & 4.3 & 4.4 & 4.0 \\
Cosmopedia-edu & 2.4 & 4.0 & 1.2 \\
Stack (code) & 2.0 & --- & --- \\
Datashop-science-qa & 1.0 & 2.1 & 0.4 \\
arXiv & --- & 2.4 & --- \\
SYNTH & --- & 4.8 & 1.0 \\
StackExchange & --- & 2.2 & 0.6 \\
Legal & --- & 0.3 & --- \\
DCLM-edu-boolq & --- & --- & 0.6 \\
\midrule
\textbf{Total} & \textbf{75.1} & \textbf{46.6} & \textbf{13.3} \\
\bottomrule
\end{tabular}
\caption{Data mixture by training stage (token counts in billions). Stage 1 emphasizes web and code data; Stage 2 adds synthetic and domain-specific sources; Stage 3 focuses on high-quality subsets with instruction formatting.}
\label{tab:data_mixture}
\end{table}

%% file: sections/eval.tex
\section{Evaluation}
\label{sec:eval}

We evaluate on standard benchmarks following the DCLM protocol~\cite{Li2024DataCompLM}: HellaSwag (commonsense reasoning), ARC-Easy/Challenge (science QA), PIQA (physical intuition), and Lambada (language modeling).
All evaluations use fixed prompting templates with deterministic decoding.
Full benchmark results and evaluation code are provided in Appendix~\ref{sec:appendix}.

%% file: sections/results.tex
\section{Results}
\label{sec:results}

\subsection{Main results}

We train \modelname{} (430M parameters) in three stages totaling 265k iterations on 72B tokens: Stage 1 (100k iterations, 29B tokens) uses standard pre-training data, Stage 2 (100k iterations, 28B tokens) applies decay with tighter quality filters, and Stage 3 (65k iterations, 14B tokens) incorporates instruction-formatted and synthetic reasoning data.
Table~\ref{tab:main_results} compares our model against SmolLM baselines.

\begin{table}[!htb]
\centering
\small
\begin{tabular}{lrrcccccc}
\toprule
\textbf{Model} & \textbf{Params} & \textbf{Tokens} & \textbf{HellaSwag} & \textbf{ARC-E} & \textbf{ARC-C} & \textbf{PIQA} & \textbf{Lambada} & \textbf{Avg} \\
\midrule
SmolLM-135M & 135M & 600B & 0.422 & 0.576 & 0.276 & 0.693 & 0.304 & 0.454 \\
SmolLM-360M & 360M & 600B & 0.540 & 0.676 & 0.351 & 0.726 & 0.509 & 0.560 \\
SmolLM2-360M & 360M & 4T & \textbf{0.554} & 0.697 & 0.411 & \textbf{0.736} & \textbf{0.532} & \textbf{0.586} \\
\midrule
\multicolumn{9}{l}{\textit{\modelname{} (430M) training progression}} \\
Stage 1 (100k) & 430M & 29B & 0.393 & 0.615 & 0.333 & 0.647 & 0.317 & 0.461 \\
Stage 2 (200k) & 430M & 57B & 0.451 & 0.689 & 0.381 & 0.682 & 0.405 & 0.522 \\
Stage 3 (265k) & 430M & 72B & 0.503 & 0.697 & 0.413 & 0.703 & 0.482 & 0.560 \\
+ EMA & 430M & 72B & 0.511 & \textbf{0.714} & \textbf{0.428} & 0.702 & 0.513 & 0.574 \\
\bottomrule
\end{tabular}
\caption{Benchmark comparison. \modelname{} trained on 72B tokens approaches SmolLM-360M (600B tokens) and SmolLM2-360M (4T tokens) performance. Bold indicates best in column.}
\label{tab:main_results}
\end{table}

\paragraph{Key findings.}
\begin{itemize}
    \item \modelname{} trained on 72B tokens approaches SmolLM-360M (600B tokens) on average benchmark score, using \textbf{8$\times$ fewer tokens}.
    \item At similar parameter count, \modelname{} approaches SmolLM2-360M (0.574 vs 0.586 avg) with \textbf{56$\times$ fewer tokens}.
    \item Three-stage training shows consistent improvements: avg 0.461 $\rightarrow$ 0.522 $\rightarrow$ 0.560, with EMA adding +0.014.
\end{itemize}

\subsection{Training efficiency}

Table~\ref{tab:compute} summarizes training compute for each stage.

\begin{table}[h]
\centering
\small
\begin{tabular}{lrrrr}
\toprule
\textbf{Stage} & \textbf{Iterations} & \textbf{Batch} & \textbf{Context} & \textbf{Wall Time (h)} \\
\midrule
Stable & 100,000 & 384 & 768 & 13.4 \\
Decay & 100,000 & 312 & 896 & 13.9 \\
Midtrain & 65,000 & 192 & 1,152 & 9.3 \\
\midrule
\textbf{Total} & 265,000 & --- & --- & \textbf{36.6} \\
\bottomrule
\end{tabular}
\caption{Training compute breakdown. Total training completed in under 37 hours on a single 8$\times$H200 node.}
\label{tab:compute}
\end{table}

Steps reported are training iterations (forward-backward passes); with gradient accumulation of 2, optimizer updates occur every 2 iterations (yielding 50k + 50k + 32.5k = 132.5k optimizer updates total).
Effective tokens per iteration: $\sim$295k (Stable), $\sim$280k (Decay), $\sim$221k (Midtrain).

\paragraph{Throughput.}
Peak throughput reaches 1.8M tokens/sec with Model FLOPs Utilization (MFU) of 26--36\% depending on stage (longer context lengths reduce MFU due to increased attention cost relative to parameter FLOPs).
NorMuon uses 7 Newton-Schulz iterations for orthogonalization~\cite{Amsel2025PolarExpress}, adding $\sim$3\% overhead compared to AdamW.

\subsection{Benchmark details}

We report accuracy on standard benchmarks following the DCLM evaluation protocol:
\begin{itemize}
    \item \textbf{CORE}: Aggregate score across held-out tasks
    \item \textbf{HellaSwag}: Commonsense reasoning (4-way)
    \item \textbf{ARC-Easy/Challenge}: Science QA
    \item \textbf{PIQA}: Physical intuition
    \item \textbf{Lambada}: Language modeling / word prediction
\end{itemize}

Additional benchmarks (Winograd, CommonsenseQA, BoolQ, etc.) are reported in Appendix~\ref{sec:appendix}.

\subsection{Training dynamics}

\begin{figure}[t]
    \centering
    \includegraphics[width=0.75\textwidth]{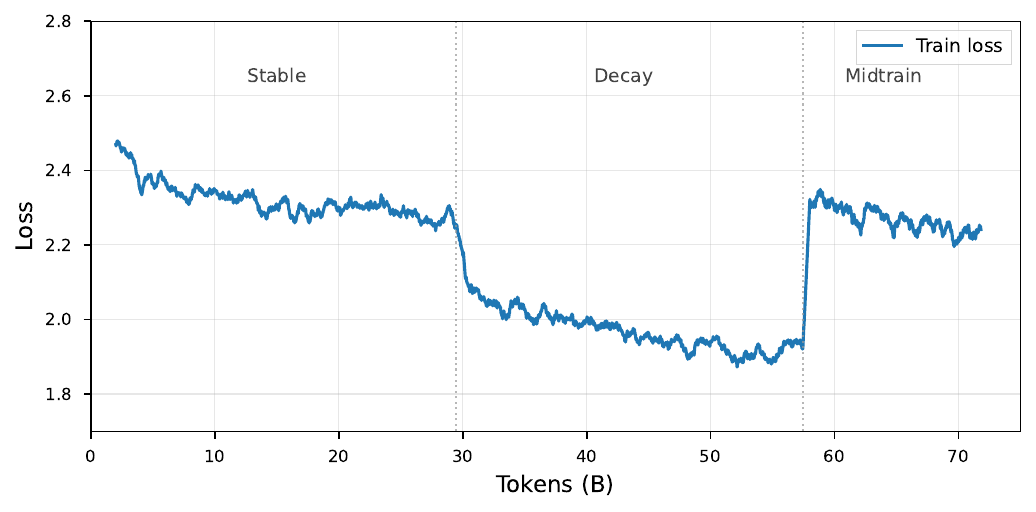}
    \caption{Training loss vs.\ tokens. Vertical lines mark stage transitions (29B, 57B).}
    \label{fig:loss_curve}
\end{figure}

Figure~\ref{fig:loss_curve} shows training loss across the three stages.

\paragraph{Combined intervention effect.}
Training the full \modelname{} architecture with all interventions disabled yields final loss 2.438; with all interventions enabled, the loss is 2.328---a 4.5\% relative improvement at matched tokens and iterations.
This validates the combined recipe on the production model, complementing the ablations on the 70M proxy model (\S\ref{sec:ablations}).

%% file: sections/ablations.tex
\section{Ablations}
\label{sec:ablations}

We conduct ablations to measure the contribution of each intervention.
All ablations use a 70M parameter model (d=768, 8 layers, 12 heads, vocab 32,768) with a custom BPE tokenizer trained on 1.4B tokens from DCLM-edu~\cite{Li2024DCLM}, representing 20$\times$ the parameter count (Chinchilla optimal~\cite{Hoffmann2022Chinchilla}).
Training uses AdamW with cosine schedule over 9,000 iterations, batch size 307,200 tokens, learning rate 0.002 with 1,000 warmup iterations and $\eta_{\min}/\eta_{\max}=0.01$.
We apply Z-loss~\cite{Chowdhery2022PaLM} with coefficient $10^{-4}$ for numerical stability.
Note: the production \modelname{} model uses a different architecture (30 layers, d=1152) and the SmolLM2-360M tokenizer (vocab 49,152); see \S\ref{sec:method} for details.

\subsection{Architectural Interventions}
\label{sec:arch_ablations}

Table~\ref{tab:arch_ablations} shows the effect of each architectural intervention in isolation.

\begin{table}[!htb]
\centering
\small
\begin{tabular}{lcccc}
\toprule
\textbf{Intervention} & \textbf{Train Loss} & \textbf{$\Delta$ Loss} & \textbf{$\Delta$ \%} \\
\midrule
Baseline (AdamW only) & 3.247 & --- & --- \\
\midrule
+ QK-Norm & 3.227 & $-$0.020 & $-$0.63\% \\
+ Per-Head Gating & 3.245 & $-$0.002 & $-$0.05\% \\
+ Norm. Value Residual & 3.234 & $-$0.013 & $-$0.38\% \\
+ LayerNorm Scaling & 3.235 & $-$0.012 & $-$0.36\% \\
\midrule
All interventions & 3.194 & $-$0.053 & $-$1.64\% \\
\bottomrule
\end{tabular}
\caption{Architectural ablations. Each row enables a single intervention versus the AdamW baseline. QK-Norm shows the largest individual benefit; the combined effect exceeds the sum of parts, suggesting positive interactions.}
\label{tab:arch_ablations}
\end{table}

QK-Norm provides the largest individual improvement ($-$0.63\%), followed by Normalized Value Residual ($-$0.38\%) and LayerNorm Scaling ($-$0.36\%). Per-Head Gating shows minimal isolated benefit.
However, all interventions combined yield $-$1.64\% improvement, exceeding the sum of individual effects.
This suggests synergistic interactions: gating may help primarily when combined with QK-Norm, and LayerNorm Scaling may become more impactful at greater depth.

\begin{figure}[!htb]
    \centering
    \includegraphics[width=\textwidth]{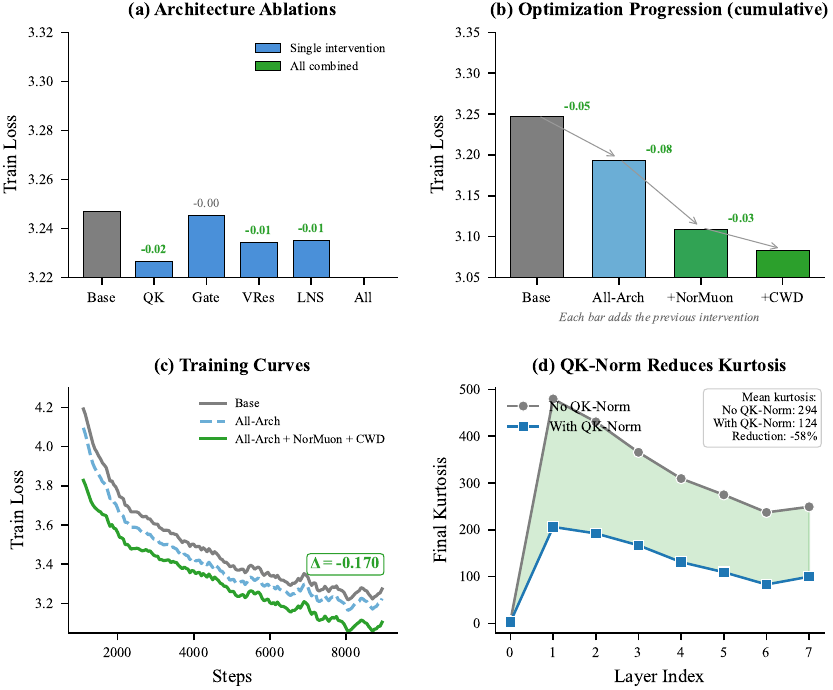}
    \caption{Ablation results. (a) Architectural interventions versus baseline. (b) Optimization progression from baseline through NorMuon with CWD. (c) Training curves comparing baseline versus all architectural interventions combined. (d) QK-Norm effect on activation kurtosis across layers---lower kurtosis indicates more stable activations.}
    \label{fig:ablations}
\end{figure}

\paragraph{QK-Norm reduces activation kurtosis.}
Figure~\ref{fig:ablations}(d) shows that QK-Norm substantially reduces kurtosis (fourth central moment) across all layers.
High kurtosis indicates heavy-tailed activation distributions that can cause training instability~\cite{Wortsman2023StableTraining}.
The kurtosis reduction is consistent across layers, supporting the stability benefits of QK-Norm.

\subsection{Optimization Interventions}
\label{sec:optim_ablations}

Building on the combined architectural interventions, we ablate optimizer choices.

\begin{table}[!htb]
\centering
\small
\begin{tabular}{lccc}
\toprule
\textbf{Configuration} & \textbf{Train Loss} & \textbf{$\Delta$ vs Arch} & \textbf{$\Delta$ vs Baseline} \\
\midrule
All Arch (AdamW) & 3.250 & --- & $-$0.047 ($-$1.42\%) \\
+ NorMuon & 3.156 & $-$0.094 ($-$2.88\%) & $-$0.141 ($-$4.27\%) \\
+ Cautious WD & 3.125 & $-$0.125 ($-$3.85\%) & $-$0.172 ($-$5.21\%) \\
\bottomrule
\end{tabular}
\caption{Optimization ablations. NorMuon provides substantial improvement over AdamW. Cautious Weight Decay adds further gains. Muon uses higher learning rate (0.0235 for 2D params, 0.007 for 1D params).}
\label{tab:optim_ablations}
\end{table}

Switching from AdamW to NorMuon yields $-$2.88\% relative improvement on the architectural baseline.
Muon tolerates higher learning rates; we use 0.0235 for 2D parameters and 0.007 for 1D parameters (embeddings, biases).
Adding Cautious Weight Decay provides an additional $-$0.97\% relative gain.

\paragraph{Total improvement.}
Combining architectural and optimization interventions yields \textbf{5.21\% relative improvement} in train loss (3.297 $\rightarrow$ 3.125) at matched tokens and iterations.

\subsection{WSD Schedule}
\label{sec:wsd_ablations}

We evaluate Warmup-Stable-Decay (WSD) schedule variants against cosine baseline.
During stable phase, we use 55\% of peak cosine LR (0.013 for Muon, 0.0039 for AdamW 1D params).

\begin{table}[!htb]
\centering
\small
\begin{tabular}{lccc}
\toprule
\textbf{Schedule} & \textbf{Train Loss} & \textbf{$\Delta$ vs Cosine} \\
\midrule
Cosine & 3.125 & --- \\
\midrule
WSD Stable only & 3.188 & +0.063 (+2.0\%) \\
WSD Decay 10\% & 3.234 & +0.109 (+3.5\%) \\
WSD Decay 20\% & \textbf{3.125} & +0.000 (0.0\%) \\
WSD Decay 30\% & 3.172 & +0.047 (+1.5\%) \\
\bottomrule
\end{tabular}
\caption{WSD schedule ablations with $1-\sqrt{t}$ decay profile. WSD with 20\% decay matches cosine performance.}
\label{tab:wsd_ablations}
\end{table}

WSD with 20\% decay matches cosine exactly (3.125), while shorter decay fractions underperform.
This suggests decay duration is critical---too little decay (10\%) leaves the model undertrained, while excessive decay (30\%) may overshoot.
We adopt WSD for production training due to its flexibility: stable-phase checkpoints can be continued with additional data before applying final decay, enabling iterative data curation without retraining from scratch.

\paragraph{Key findings.}
\begin{itemize}
    \item Architectural interventions provide 1.42\% relative improvement, with QK-Norm contributing the most individually.
    \item NorMuon with Cautious WD provides an additional 3.85\% relative improvement over AdamW.
    \item Combined interventions yield \textbf{5.21\% relative improvement} in train loss at matched tokens and iterations.
    \item QK-Norm measurably reduces activation kurtosis across all layers.
\end{itemize}

%% file: sections/repro.tex
\section{Release and Reproducibility}
\label{sec:repro}

We release (i) model weights, (ii) tokenizer, (iii) training and evaluation code, and (iv) exact configuration files and run manifests.
Training and evaluation are implemented in \texttt{sample\_efficient\_gpt} \cite{Deez2026SampleEfficientGPT}.
We additionally provide inference wrappers (Hugging Face / vLLM) and scripts to reproduce all figures and tables from logged metrics.

\paragraph{Artifacts.}
\begin{itemize}
  \item \textbf{Model:} \url{https://huggingface.co/thepowerfuldeez/imu1_base} (base checkpoint with EMA).
  \item \textbf{Code:} \url{https://github.com/thepowerfuldeez/sample_efficient_gpt}.
  \item \textbf{Data:} \url{https://huggingface.co/datasets/thepowerfuldeez/1218_imu1_base_stable_corpus}, \url{https://huggingface.co/datasets/thepowerfuldeez/1226_imu1_base_decay_corpus}.
\end{itemize}

%% file: sections/limitations.tex
\section{Limitations}
\label{sec:limitations}

Our experiments focus on sub-billion parameter models (440M) and a specific data mixture. While \muP{} parametrization is designed for hyperparameter transfer, we have not validated that all interventions provide consistent benefits at larger scales or with different data compositions.

%% file: sections/appendix.tex
\section{Additional Results}
\label{sec:appendix}

\subsection{Full benchmark scores}

Table~\ref{tab:full_benchmarks} provides complete benchmark results for \modelname{} at each training stage.

\begin{table}[h]
\centering
\scriptsize
\begin{tabular}{lcccc}
\toprule
\textbf{Benchmark} & \textbf{100k (stable)} & \textbf{200k (decay)} & \textbf{265k (midtrain)} & \textbf{EMA} \\
\midrule
HellaSwag (0-shot) & 0.393 & 0.451 & 0.503 & 0.511 \\
Jeopardy & 0.128 & 0.224 & 0.268 & 0.288 \\
BigBench QA WikiData & 0.516 & 0.560 & 0.582 & 0.599 \\
ARC-Easy & 0.615 & 0.689 & 0.697 & 0.714 \\
ARC-Challenge & 0.333 & 0.381 & 0.413 & 0.411 \\
COPA & 0.630 & 0.720 & 0.700 & 0.690 \\
CommonsenseQA & 0.260 & 0.400 & 0.421 & 0.430 \\
PIQA & 0.647 & 0.682 & 0.703 & 0.702 \\
OpenBookQA & 0.342 & 0.372 & 0.386 & 0.382 \\
Lambada (OpenAI) & 0.317 & 0.405 & 0.482 & 0.513 \\
Winograd & 0.659 & 0.674 & 0.722 & 0.747 \\
WinoGrande & 0.523 & 0.552 & 0.547 & 0.552 \\
BigBench Dyck Languages & 0.198 & 0.144 & 0.147 & 0.162 \\
AGI Eval LSAT AR & 0.252 & 0.274 & 0.235 & 0.261 \\
BigBench CS Algorithms & 0.395 & 0.455 & 0.437 & 0.439 \\
BigBench Operators & 0.243 & 0.300 & 0.271 & 0.338 \\
BigBench Repeat Copy Logic & 0.031 & 0.063 & 0.063 & 0.063 \\
SQuAD & 0.094 & 0.224 & 0.263 & 0.319 \\
CoQA & 0.209 & 0.288 & 0.316 & 0.353 \\
BoolQ & 0.532 & 0.460 & 0.520 & 0.595 \\
BigBench Language ID & 0.250 & 0.255 & 0.264 & 0.269 \\
\midrule
\textbf{CORE (centered)} & 0.192 & 0.249 & 0.276 & \textbf{0.302} \\
\bottomrule
\end{tabular}
\caption{Full benchmark results for \modelname{} across training stages.}
\label{tab:full_benchmarks}
\end{table}

For context, Granite-4.0-350M Dense~\cite{Granite42025} reports AGIEval aggregate of 28.97 (20 tasks) and BBH aggregate of 32.19 (23 tasks with CoT); our DCLM evaluation includes LSAT-AR (26.09) and Dyck Languages (16.20) as individual tasks from these suites. Granite uses 14.5T tokens (200$\times$ more than our 72B).

\subsection{Model architecture}

\begin{table}[h]
\centering
\small
\begin{tabular}{ll}
\toprule
\textbf{Parameter} & \textbf{Value} \\
\midrule
Hidden dimension ($d_{model}$) & 1,152 \\
Number of layers & 30 \\
Number of heads & 18 \\
Head dimension & 64 \\
FFN hidden dimension & 3,072 \\
Vocabulary size & 49,152 \\
KV heads (GQA) & 6 \\
Total parameters & $\sim$440M \\
\midrule
\multicolumn{2}{l}{\textit{Architectural features}} \\
QK-norm & Yes (learnable scale) \\
Per-head gating & Yes (sigmoid) \\
Value residual & Yes \\
LayerNorm scaling & Yes (depth-dependent) \\
Position encoding & RoPE ($\theta=10000$) \\
Activation & SwiGLU \\
\bottomrule
\end{tabular}
\caption{Model architecture details.}
\label{tab:architecture}
\end{table}

\subsection{Hyperparameters}
\label{sec:hyperparams}

Table~\ref{tab:hparams} lists key hyperparameters for each training stage.

\begin{table}[h]
\centering
\small
\begin{tabular}{llll}
\toprule
\textbf{Parameter} & \textbf{Stable} & \textbf{Decay} & \textbf{Midtrain} \\
\midrule
Steps & 100,000 & 100,000 & 65,000 \\
Batch size & 384 & 312 & 192 \\
Context length & 768 & 896 & 1,152 \\
Muon LR & 0.011 & 0.0115 & 0.003 \\
Embed LR & 0.006 & 0.006 & 0.002 \\
Muon WD & 0.1 & 0.1 & 0.1 \\
LR min ratio & 0.01 & 0.25 & 0.33 \\
Warmup steps & 2,500 & 0 & 0 \\
Grad accum & 2 & 2 & 2 \\
Z-loss weight~\cite{Chowdhery2022PaLM} & $10^{-4}$ & $10^{-4}$ & $10^{-4}$ \\
\bottomrule
\end{tabular}
\caption{Training hyperparameters by stage.}
\label{tab:hparams}
\end{table}

\paragraph{Step and token accounting.}
``Steps'' refers to training iterations (forward-backward passes), not optimizer steps.
With gradient accumulation of 2, optimizer updates occur every 2 iterations.
Effective tokens per iteration: Stable = $384 \times 768 = 295$k, Decay = $312 \times 896 = 280$k, Midtrain = $192 \times 1152 = 221$k.
Total tokens: $100\text{k} \times 295\text{k} + 100\text{k} \times 280\text{k} + 65\text{k} \times 221\text{k} \approx 72$B.

\paragraph{Checkpoint EMA.}
We apply exponential moving average to checkpoints post-hoc with $\beta = 0.8$ over the final 10 checkpoints (saved every 5k steps during decay phase).
The reported ``+ EMA'' results use this averaged checkpoint.

\subsection{WSD schedule comparison}

Each stage uses Warmup-Stable-Decay (WSD) with $1-\sqrt{t}$ decay profile.
The stable phase uses 55\% of peak cosine LR.
Decay begins at 80\% of total stage steps (i.e., 20\% decay fraction based on ablations).

\begin{table}[!htb]
\centering
\small
\begin{tabular}{lccc}
\toprule
\textbf{Schedule} & \textbf{Decay \%} & \textbf{Steps} & \textbf{Final Loss} \\
\midrule
Cosine & 100\% & 9,000 & \textbf{3.125} \\
WSD & 0\% (stable only) & 9,000 & 3.188 \\
WSD & 10\% & 9,000 & 3.234 \\
WSD & 20\% & 9,000 & \textbf{3.125} \\
WSD & 30\% & 9,000 & 3.172 \\
\bottomrule
\end{tabular}
\caption{WSD schedule comparison on 70M model with NorMuon+CWD. WSD with 20\% decay matches cosine performance while enabling checkpoint re-use for multi-stage training.}
\label{tab:wsd_ablation}
\end{table}

\subsection{DCLM-edu-boolq filtering prompt}
\label{sec:boolq_prompt}

The following prompt identifies passages suitable for Boolean question answering.
An LLM processes batches of 50 samples and returns indices of qualifying passages; these are used to train a fastText classifier.

\begin{quote}
\small
\texttt{You are selecting text samples suitable for training Boolean question answering and textual entailment models (BoolQ / NLI style).}

\texttt{A sample qualifies ONLY IF ALL conditions hold:}

\texttt{1) The passage contains at least one explicit or implicit factual CLAIM whose truth value can be determined from the passage itself.}

\texttt{2) That claim could naturally be turned into a YES/NO question (e.g., involving negation, exceptions, quantifiers, or exclusivity).}

\texttt{3) The passage provides sufficient EVIDENCE to verify or falsify the claim. (No external knowledge required.)}

\texttt{4) Preference is given to passages involving: negation (not, no, never, except, unless); quantifiers (all, some, only, at least, none); contradictions or contrastive statements; factual verification.}

\texttt{5) EXCLUDE passages that are: purely descriptive or narrative; definitions without verifiable truth conditions; open-ended discussion without decidable claims; opinion or speculative text.}

\texttt{If NO samples qualify, return an EMPTY LINE. Output ONLY the numbers of qualifying samples as a comma-separated list.}
\end{quote}

%% file: main.bbl
\begin{thebibliography}{34}
\providecommand{\natexlab}[1]{#1}
\providecommand{\url}[1]{\texttt{#1}}
\expandafter\ifx\csname urlstyle\endcsname\relax
  \providecommand{\doi}[1]{doi: #1}\else
  \providecommand{\doi}{doi: \begingroup \urlstyle{rm}\Url}\fi

\bibitem[Ahn et~al.(2025)Ahn, Xu, Abreu, Fan, Magakyan, Sharma, Zhan, and
  Langford]{Ahn2025Dion}
Kwangjun Ahn, Byron Xu, Natalie Abreu, Ying Fan, Gagik Magakyan, Pratyusha
  Sharma, Zheng Zhan, and John Langford.
\newblock Dion: Distributed orthonormalized updates, 2025.

\bibitem[Ainslie et~al.(2023)Ainslie, Lee-Thorp, de~Jong, Zemlyanskiy, Lebron,
  and Sanghai]{Ainslie2023GQA}
Joshua Ainslie, James Lee-Thorp, Michiel de~Jong, Yury Zemlyanskiy, Federico
  Lebron, and Sumit Sanghai.
\newblock Gqa: Training generalized multi-query transformer models from
  multi-head checkpoints, 2023.

\bibitem[Allal et~al.(2025)Allal, Lozhkov, Bakouch, Bl{\'a}zquez, Penedo,
  Tunstall, et~al.]{BenAllal2025SmolLM2}
Loubna~Ben Allal, Anton Lozhkov, Elie Bakouch, Gabriel~Mart{\'\i}n
  Bl{\'a}zquez, Guilherme Penedo, Lewis Tunstall, et~al.
\newblock Smollm2: When smol goes big -- data-centric training of a small
  language model, 2025.

\bibitem[Amsel et~al.(2025)Amsel, Persson, Musco, and
  Gower]{Amsel2025PolarExpress}
Noah Amsel, David Persson, Christopher Musco, and Robert~M. Gower.
\newblock The polar express: Optimal matrix sign methods and their application
  to the muon algorithm, 2025.

\bibitem[Chen et~al.(2025)Chen, Li, Liang, Su, Xie, Pierse, Liang, Lao, and
  Liu]{Chen2025CWD}
Lizhang Chen, Jonathan Li, Kaizhao Liang, Baiyu Su, Cong Xie, Nuo~Wang Pierse,
  Chen Liang, Ni~Lao, and Qiang Liu.
\newblock Cautious weight decay, 2025.

\bibitem[Chowdhery et~al.(2022)Chowdhery, Narang, Devlin, Bosma, Mishra,
  Roberts, et~al.]{Chowdhery2022PaLM}
Aakanksha Chowdhery, Sharan Narang, Jacob Devlin, Maarten Bosma, Gaurav Mishra,
  Adam Roberts, et~al.
\newblock Palm: Scaling language modeling with pathways, 2022.
\newblock Z-loss regularization.

\bibitem[Dao et~al.(2022)Dao, Fu, Ermon, Rudra, and
  R{\'e}]{Dao2022FlashAttention}
Tri Dao, Daniel~Y. Fu, Stefano Ermon, Atri Rudra, and Christopher R{\'e}.
\newblock Flashattention: Fast and memory-efficient exact attention with
  io-awareness, 2022.

\bibitem[Dehghani et~al.(2023)Dehghani, Djolonga, Mustafa, Padlewski, Heek,
  Gilmer, et~al.]{Dehghani2023ScalingViT}
Mostafa Dehghani, Josip Djolonga, Basil Mustafa, Piotr Padlewski, Jonathan
  Heek, Justin Gilmer, et~al.
\newblock Scaling vision transformers to 22 billion parameters, 2023.
\newblock QK-Norm for ViT stability.

\bibitem[{Gemma Team}(2025)]{Gemma32025}
{Gemma Team}.
\newblock Gemma 3 technical report, 2025.
\newblock URL \url{https://ai.google.dev/gemma/docs/core}.
\newblock Google DeepMind.

\bibitem[Henry et~al.(2020)Henry, Dachapally, Pawar, and Chen]{Henry2020QKNorm}
Alex Henry, Prudhvi~Raj Dachapally, Shubham Pawar, and Yuxuan Chen.
\newblock Query-key normalization for transformers, 2020.

\bibitem[Hoffmann et~al.(2022)Hoffmann, Borgeaud, Mensch, Buchatskaya, Cai,
  Rutherford, et~al.]{Hoffmann2022Chinchilla}
Jordan Hoffmann, Sebastian Borgeaud, Arthur Mensch, Elena Buchatskaya, Trevor
  Cai, Eliza Rutherford, et~al.
\newblock Training compute-optimal large language models, 2022.

\bibitem[{IBM Research}(2025)]{Granite42025}
{IBM Research}.
\newblock Granite 4.0: Hyper-efficient, high performance hybrid models for
  enterprise, 2025.
\newblock URL \url{https://huggingface.co/ibm-granite/granite-4.0-h-350m-base}.

\bibitem[Li et~al.(2024{\natexlab{a}})Li, Fang, Smyrnis, Ivgi, Jordan, Gadre,
  et~al.]{Li2024DCLM}
Jeffrey Li, Alex Fang, Georgios Smyrnis, Maor Ivgi, Matt Jordan, Samir Gadre,
  et~al.
\newblock Datacomp-lm: In search of the next generation of training sets for
  language models, 2024{\natexlab{a}}.

\bibitem[Li et~al.(2024{\natexlab{b}})Li, Fang, Smyrnis, Ivgi, Jordan, Gadre,
  et~al.]{Li2024DataCompLM}
Jeffrey Li, Alex Fang, Georgios Smyrnis, Maor Ivgi, Matt Jordan, Samir Gadre,
  et~al.
\newblock Datacomp-lm: In search of the next generation of training sets for
  language models, 2024{\natexlab{b}}.

\bibitem[Li et~al.(2025)Li, Liu, Liang, Chen, and Zhao]{Li2025NorMuon}
Zichong Li, Liming Liu, Chen Liang, Weizhu Chen, and Tuo Zhao.
\newblock Normuon: Making muon more efficient and scalable, 2025.

\bibitem[{Liquid AI}(2025)]{LFM22025}
{Liquid AI}.
\newblock Lfm2 technical report, 2025.

\bibitem[Loshchilov and Hutter(2017)]{Loshchilov2017SGDR}
Ilya Loshchilov and Frank Hutter.
\newblock Sgdr: Stochastic gradient descent with warm restarts.
\newblock In \emph{International Conference on Learning Representations}, 2017.
\newblock URL \url{https://openreview.net/forum?id=Skq89Scxx}.

\bibitem[Luo et~al.(2025)Luo, Sun, Wen, Shi, Cui, Dang, Lyu, and
  Chen]{Luo2025LRDecayCurriculum}
Kairong Luo, Zhenbo Sun, Haodong Wen, Xinyu Shi, Jiarui Cui, Chenyi Dang,
  Kaifeng Lyu, and Wenguang Chen.
\newblock How learning rate decay wastes your best data in curriculum-based llm
  pretraining, 2025.

\bibitem[Penedo et~al.(2023)Penedo, Malartic, Hesslow, Cojocaru, Cappelli,
  Alobeidli, Pannier, Almazrouei, and Launay]{Penedo2023RefinedWeb}
Guilherme Penedo, Quentin Malartic, Daniel Hesslow, Ruxandra Cojocaru,
  Alessandro Cappelli, Hamza Alobeidli, Baptiste Pannier, Ebtesam Almazrouei,
  and Julien Launay.
\newblock The refinedweb dataset for falcon llm: Outperforming curated corpora
  with web data, and web data only, 2023.

\bibitem[{Pleias}(2024)]{Pleias2024Blogsynth}
{Pleias}.
\newblock Synth: the new data frontier, 2024.
\newblock URL \url{https://pleias.fr/blog/blogsynth-the-new-data-frontier}.
\newblock Blog post.

\bibitem[{Pleias}(2025)]{Baguettotron2025}
{Pleias}.
\newblock Baguettotron: A 321m reasoning model, 2025.
\newblock URL \url{https://huggingface.co/PleIAs/Baguettotron}.
\newblock Trained on SYNTH dataset.

\bibitem[Qiu et~al.(2025)Qiu, Wang, Zheng, Huang, Wen, Yang, Men, Yu, Huang,
  Huang, Liu, Zhou, and Lin]{Qiu2025GatedAttention}
Zihan Qiu, Zekun Wang, Bo~Zheng, Zeyu Huang, Kaiyue Wen, Songlin Yang, Rui Men,
  Le~Yu, Fei Huang, Suozhi Huang, Dayiheng Liu, Jingren Zhou, and Junyang Lin.
\newblock Gated attention for large language models: Non-linearity, sparsity,
  and attention-sink-free, 2025.

\bibitem[Shah et~al.(2025)Shah, Polloreno, Stratos, Monk, Chaluvaraju, Hojel,
  Ma, Thomas, Tanwer, Shah, Nguyen, Smith, Callahan, Pust, Parmar, Rushton,
  Mazarakis, Kapila, Srivastava, Singla, Romanski, Vanjani, and
  Vaswani]{EssentialAI2025Muon}
Ishaan Shah, Anthony~M. Polloreno, Karl Stratos, Philip Monk, Adarsh
  Chaluvaraju, Andrew Hojel, Andrew Ma, Anil Thomas, Ashish Tanwer, Darsh~J
  Shah, Khoi Nguyen, Kurt Smith, Michael Callahan, Michael Pust, Mohit Parmar,
  Peter Rushton, Platon Mazarakis, Ritvik Kapila, Saurabh Srivastava, Somanshu
  Singla, Tim Romanski, Yash Vanjani, and Ashish Vaswani.
\newblock Practical efficiency of muon for pretraining, 2025.

\bibitem[Shazeer(2020)]{Shazeer2020GLU}
Noam Shazeer.
\newblock Glu variants improve transformer, 2020.

\bibitem[Su et~al.(2021)Su, Lu, Pan, Murtadha, Wen, and Liu]{Su2021RoPE}
Jianlin Su, Yu~Lu, Shengfeng Pan, Ahmed Murtadha, Bo~Wen, and Yunfeng Liu.
\newblock Roformer: Enhanced transformer with rotary position embedding, 2021.

\bibitem[Sun et~al.(2025)Sun, Song, Li, Yin, Zheng, and Liu]{Sun2025CurseDepth}
Wenfang Sun, Xinyuan Song, Pengxiang Li, Lu~Yin, Yefeng Zheng, and Shiwei Liu.
\newblock The curse of depth in large language models, 2025.

\bibitem[(thepowerfuldeez)(2026)]{Deez2026SampleEfficientGPT}
George~Grigorev (thepowerfuldeez).
\newblock sample\_efficient\_gpt: Training framework for sample-efficient llm
  pretraining, 2026.
\newblock URL \url{https://github.com/thepowerfuldeez/sample_efficient_gpt}.

\bibitem[Tian et~al.(2025)Tian, Wang, Zhao, Chen, Liu, Liu, Mao, Zhao, Zhang,
  and Zhou]{Tian2025WSM}
Changxin Tian, Jiapeng Wang, Qian Zhao, Kunlong Chen, Jia Liu, Ziqi Liu, Jiaxin
  Mao, Wayne~Xin Zhao, Zhiqiang Zhang, and Jun Zhou.
\newblock Wsm: Decay-free learning rate schedule via checkpoint merging for llm
  pre-training, 2025.

\bibitem[Wen et~al.(2024)Wen, Li, Wang, Hall, Liang, and Ma]{Wen2024WSD}
Kaiyue Wen, Zhiyuan Li, Jason Wang, David Hall, Percy Liang, and Tengyu Ma.
\newblock Understanding warmup-stable-decay learning rates: A river valley loss
  landscape perspective, 2024.

\bibitem[Wortsman et~al.(2023)Wortsman, Liu, Xiao, Everett, Alemi, Adlam,
  Co-Reyes, Gur, Kumar, Novak, Pennington, Sohl-Dickstein, Xu, Lee, Gilmer, and
  Kornblith]{Wortsman2023StableTraining}
Mitchell Wortsman, Peter~J. Liu, Lechao Xiao, Katie Everett, Alex Alemi, Ben
  Adlam, John~D. Co-Reyes, Izzeddin Gur, Abhishek Kumar, Roman Novak, Jeffrey
  Pennington, Jascha Sohl-Dickstein, Kelvin Xu, Jaehoon Lee, Justin Gilmer, and
  Simon Kornblith.
\newblock Stable training of large language models with adaptive gradient
  methods, 2023.

\bibitem[Xiao et~al.(2023)Xiao, Tian, Chen, Han, and
  Lewis]{Xiao2023StreamingLLM}
Guangxuan Xiao, Yuandong Tian, Beidi Chen, Song Han, and Mike Lewis.
\newblock Efficient streaming language models with attention sinks, 2023.

\bibitem[Yang et~al.(2022)Yang, Hu, Babuschkin, Sidor, Liu, Farhi, Ryder,
  Pachocki, Chen, and Gao]{Yang2022TensorProgramsV}
Greg Yang, Edward~J. Hu, Igor Babuschkin, Szymon Sidor, Xiaodong Liu, David
  Farhi, Nick Ryder, Jakub Pachocki, Weizhu Chen, and Jianfeng Gao.
\newblock Tensor programs v: Tuning large neural networks via zero-shot
  hyperparameter transfer, 2022.

\bibitem[Zhang and Sennrich(2019)]{Zhang2019RMSNorm}
Biao Zhang and Rico Sennrich.
\newblock Root mean square layer normalization, 2019.

\bibitem[Zhou et~al.(2024)Zhou, Wu, Jiang, Obeid, and
  Lan]{Zhou2024ValueResidual}
Zhanchao Zhou, Tianyi Wu, Zhiyun Jiang, Fares Obeid, and Zhenzhong Lan.
\newblock Value residual learning, 2024.

\end{thebibliography}
